\documentclass[final]{cvpr}

\usepackage{times}
\usepackage{epsfig}
\usepackage{graphicx}
\usepackage{amsmath}
\usepackage{amssymb}
\usepackage{caption}
\usepackage[pagebackref=true,breaklinks=true,colorlinks,bookmarks=false]{hyperref}

\def\cvprPaperID{3639} 
\def\confYear{CVPR 2021}

\begin{document}

\title{ 
Pushing it out of the Way: Interactive Visual Navigation }

\author{Kuo-Hao Zeng$^{1}$\ \ \ \ \ \ Luca Weihs$^{2}$\ \ \ \ \ \ Ali Farhadi$^{1}$\ \ \ \ \ \ Roozbeh Mottaghi$^{1,2}$\\
\normalsize{$^{1}$Paul G. Allen School of Computer Science \& Engineering, University of Washington}\\
\normalsize{$^{2}$PRIOR @ Allen Institute for AI}\\
\url{prior.allenai.org/projects/interactive-visual-navigation}
}

\twocolumn[{
\renewcommand\twocolumn[1][]{#1}
\maketitle
\centering
\includegraphics[width=.99\linewidth]{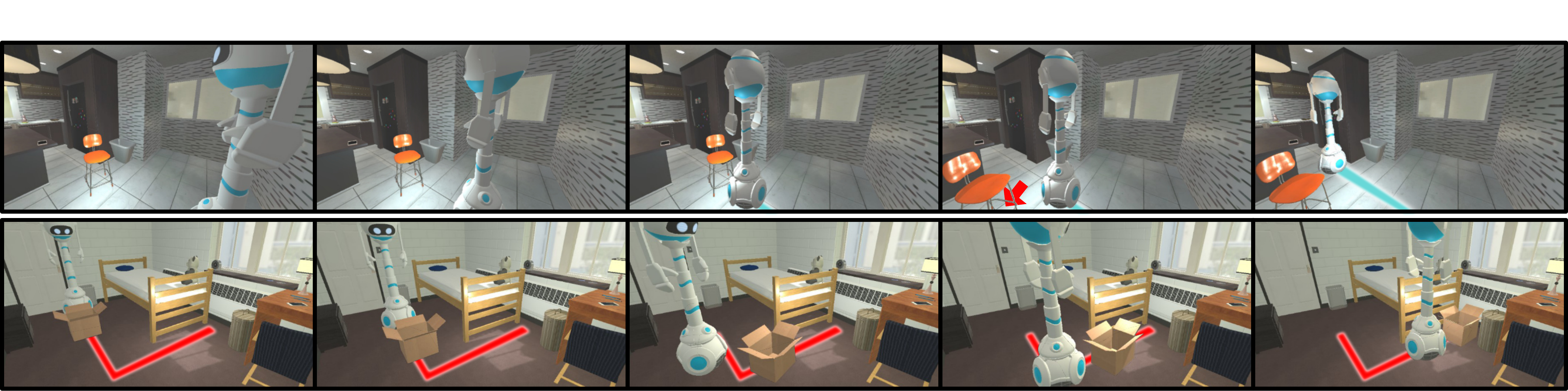}
\vspace{-.2cm}
\captionof{figure}{
Visual navigation may require interactions that go beyond moving forward/backward, and turning left/right. For example, the agent in the top row needs to push the chair out of its way to reach the target. Interactive navigation entails deeper understanding of the outcome of agents actions on objects in the scene. In this paper, we introduce Neural Interaction Engine (NIE) to explicitly predict the effect of actions on objects poses. By integrating NIE with our policy network we show that we can perform long-horizon planning while predicting the outcome of the actions. We evaluate NIE for visual navigation where the path to the goal is obstructed, and moving objects to specific locations in the scene and show major improvements over state of the art in these tasks. 
}
\label{fig:teaser}
\vspace*{0.25cm}
}]

\maketitle
\thispagestyle{empty}

\begin{abstract}
\vspace*{-0.39cm}
We have observed significant progress in visual navigation for embodied agents. A common assumption in studying visual navigation is that the environments are static; this is a limiting assumption. Intelligent navigation may involve interacting with the environment beyond just moving forward/backward and turning left/right. Sometimes, the best way to navigate is to push something out of the way. In this paper, we study the problem of interactive navigation where agents learn to change the environment to navigate more efficiently to their goals. To this end, we introduce the Neural Interaction Engine (NIE) to explicitly predict the change in the environment caused by the agent's actions. By modeling the changes while planning, we find that agents exhibit significant improvements in their navigational capabilities. More specifically, we consider two downstream tasks in the physics-enabled, visually rich, AI2-THOR environment: (1) reaching a target while the path to the target is blocked (2) moving an object to a target location by pushing it. For both tasks, agents equipped with an NIE significantly outperform agents without the understanding of the effect of the actions indicating the benefits of our approach.

\end{abstract}


\section{Introduction}

Embodied AI has witnessed remarkable progress over the past few years owing to advances in learning algorithms, benchmarks, and standardized tasks. A popular task that has received a considerable amount of attention is visual navigation \cite{anderson18,Batra2020ObjectNavRO,chaplot2020object,Savva_2019_ICCV,Wortsman2019LearningTL,zhu17}, where the goal is to navigate towards a specific coordinate or object within an unseen environment. One of the common implicit assumptions for these navigation methods is that the scene is static, and the agent cannot interact with the objects to change their pose.

Consider the scenario that the path of the agent towards the target location is blocked by an obstacle (e.g., a chair) as shown in Fig.~\ref{fig:teaser}~(top). To reach the target, the agent has to move the obstacle out of the way. Therefore, planning for reaching the target requires not only understanding the outcome of agent actions but also the dynamics of agent-object interactions. There are many factors such as object size, spatial relationship with other objects in the scene, and reaction of the object to the applied forces, that influence the outcome of the interaction with the object. Hence, long-horizon planning for navigation conditioned on the object dynamics offers unique challenges that are often overlooked in the recent navigation literature.

The first challenge is to learn whether an action affects the pose of an object or not. Navigation actions (e.g., rotate right or move ahead) typically do not affect the position of objects in the world coordinate frame while interaction actions (e.g., pushing an object) can change the object pose. The objects move in the ego-centric view of the agent due to agent movements or interaction with objects. Learning how objects move as a result of camera motion or interaction imposes the second challenge. Learning how to interact with objects is another challenge. For example, the agent should learn that pushing an object against a wall does not change its pose.

In this paper, we propose a novel model for navigation while interacting  with objects within a scene that jointly plans a sequence of actions and predicts the changes in the scene conditioned on those actions. More specifically, the model includes a Neural Interaction Engine (NIE) module that predicts the affine transformation of objects from the perspective of the agent conditioned on the actions. The goal is to learn if/how the actions affect the pose of the objects. The NIE module receives gradients for not only the prediction of the pose in the next frame but also the navigation policy.

We evaluate our model on two downstream tasks \emph{ObsNav} and \emph{ObjPlace}. The goal of \emph{ObsNav} is to reach a specific coordinates in a scene while the paths from the initial location of the agent to the target are blocked by objects. The goal of \emph{ObjPlace} is to push an object on the floor while navigating so it reaches a target point. These are challenging tasks since the agent requires an accurate understanding of the dynamics of the objects and their interaction with other objects in the scene. We perform our experiments in 120 scenes of the physics-enabled AI2-THOR \cite{kolve2017ai2} environment. Our experiments show significant improvement over baselines that are not capable of explicitly predicting the effect of interactions showing the merit of our NIE model. 

In summary, we highlight three primary contributions. (1) We propose Neural Interaction Engine, as a model for predicting the state of the observed objects conditioned on the agent actions. (2) We propose new datasets for two navigation-based tasks using a physics-enabled framework, which enables changing the pose of objects and models rich object-object and agent-object interactions. (3) We show that predicting the outcome of actions is a crucial capability for embodied agents by showing significant improvements over baselines that do not possess this capability.

\section{Related Work}
\noindent \textbf{Action-conditioned learning of rigid body dynamics.} The goal of these works is to learn the dynamics of rigid body motion under the effect of applied actions. Byravan and Fox~\cite{Byravan2017SE3netsLR} segment a point cloud into salient regions and predict the rigid body motion. Li \emph{et al}.~\cite{Li-RSS-18} learn to re-position and re-orient an object with unknown physical properties. Several works~\cite{foresight,Ebert2017SelfSupervisedVP,Finn2017DeepVF,zeng2020visual} have proposed formulations of visual Model Predictive Control, where the central insight is that a predictive model of sensory input is a powerful signal for learning to perform tasks. A number of other strategies for action-conditioned learning have been proposed, these include: learning latent physical properties of objects using visual observation of interactions with those objects~\cite{Xu2019DensePhysNetLD}, learning forward and inverse scene dynamics from object interaction data~\cite{nematoli20iros}, representing scenes as object-centric graphs and learning to predict changes in object pose after applying a push action~\cite{Paus2020PredictingPA}, learning the dynamics of balls and walls in the game of billiards~\cite{billiards}, and modeling the dynamics of robot interactions by jointly estimating forward and inverse models of dynamics~\cite{agrawal16}. In contrast to all of these approaches, we consider the more complex mobile robot scenario, where we factorize the effect of robot motion and object motion. 

\noindent \textbf{Learning dynamics from perception.} The dynamics of objects can be inferred from images and videos alone without any interaction. \cite{sfmnet} decompose frame-to-frame pixel motion into scene depth, 3D camera rotation and translation, and a set of object regions with their corresponding 3D motion. \cite{li2020visual} reason about the underlying physical properties of objects that appear in a sequence of frames and predict future motion of those objects. \cite{janner2019reasoning,deanim} jointly train a perception module, an object-based physics engine and a renderer to generate the future predictions. \cite{chang2016compositional} propose Neural Physics Engine that outputs the future states of objects and their properties. \cite{interactionnet} also infers the physical state of objects from video input and predict their future trajectories. \cite{phys101} infer physical properties of objects such as mass and density from videos. \cite{mottaghi16,mottaghi16b,zeng2020visual,zeng2017visual} predict the dynamics of objects and their future trajectory. These approaches focus on simple scenarios (such as balls of uniform mass or a stack of cubes), no agent action is considered or assume a static camera. 

\noindent \textbf{Visual navigation.} The tasks that we consider in this paper involves visual navigation. Visual navigation has been addressed in various papers in recent Embodied AI literature. Most works focus on point navigation (PointNav) \cite{anderson18,chaplot2020learning,Savva_2019_ICCV,Wijmans2020DDPPOLN} or object navigation (ObjectNav) \cite{Batra2020ObjectNavRO,chaplot2020object,du20,Wortsman2019LearningTL}. Our task is different since in these works only static scenes are considered. 

Our task is closer to existing tasks that consider navigation among movable obstacles~\cite{batra2020rearrangement,karkus2020beyond,mirza2020physically,Stilman-2004-9082,Stilman2008PlanningAM,xiali2020relmogen,xia2020interactive}. The difference with \cite{Stilman-2004-9082,Stilman2008PlanningAM} is that those works are not learning-based and generalization to unseen scenes is not evaluated. Our task differs from that of \cite{xia2020interactive} in that our agent applies forces to objects with different magnitudes and directions (as opposed to moving objects by colliding with them). Our approach also shows significant improvements over the vanilla RL approaches used in \cite{xia2020interactive}.


\section{Model}

\begin{figure}[tp]
\includegraphics[width=19pc]{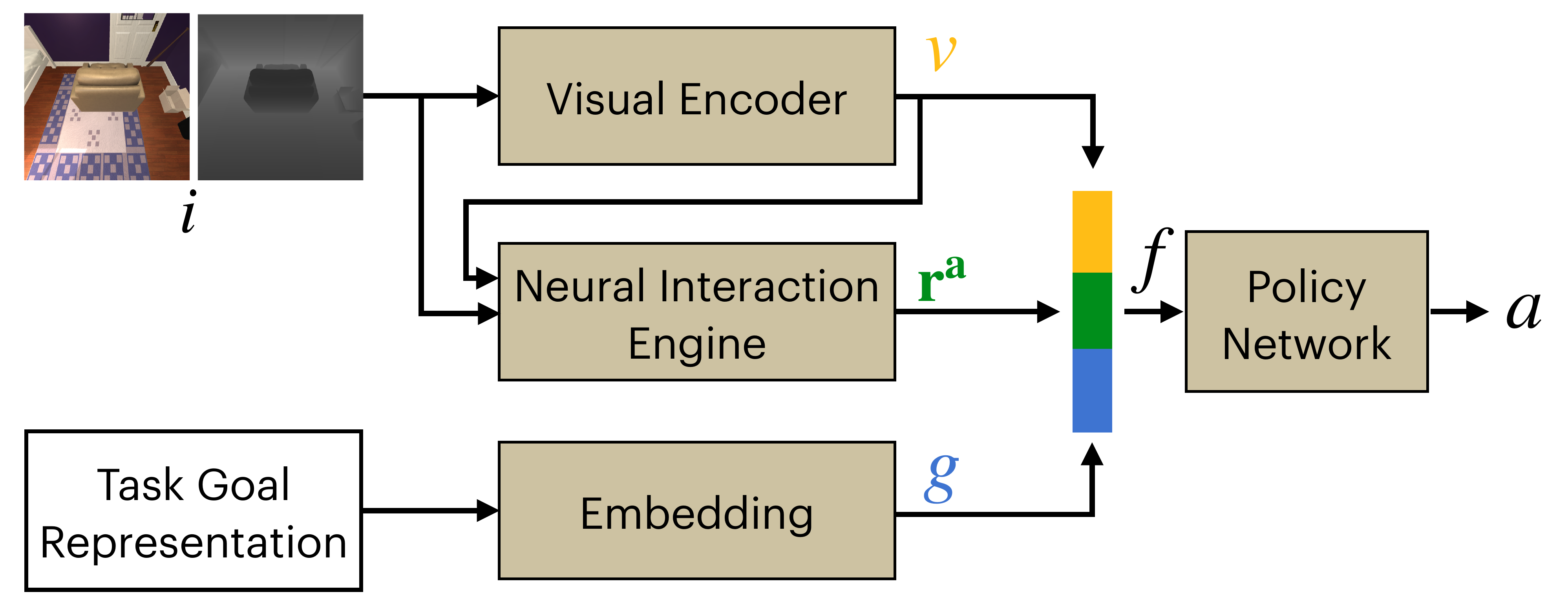}
\caption{\textbf{Model overview.} Our model includes three main parts: Visual Encoder, Neural Interaction Engine, and Policy Network.} 
\label{fig:framework}
\end{figure}

In this section, we begin by providing an overview of the proposed model. We then introduce our Neural Interaction Engine (NIE) and explain how we integrate the NIE into the policy network. Finally, we describe the learning objective and how we learn the entire model with the NIE module.

\subsection{Model Overview}

\begin{figure}[bp]
\centering
\includegraphics[width=19pc]{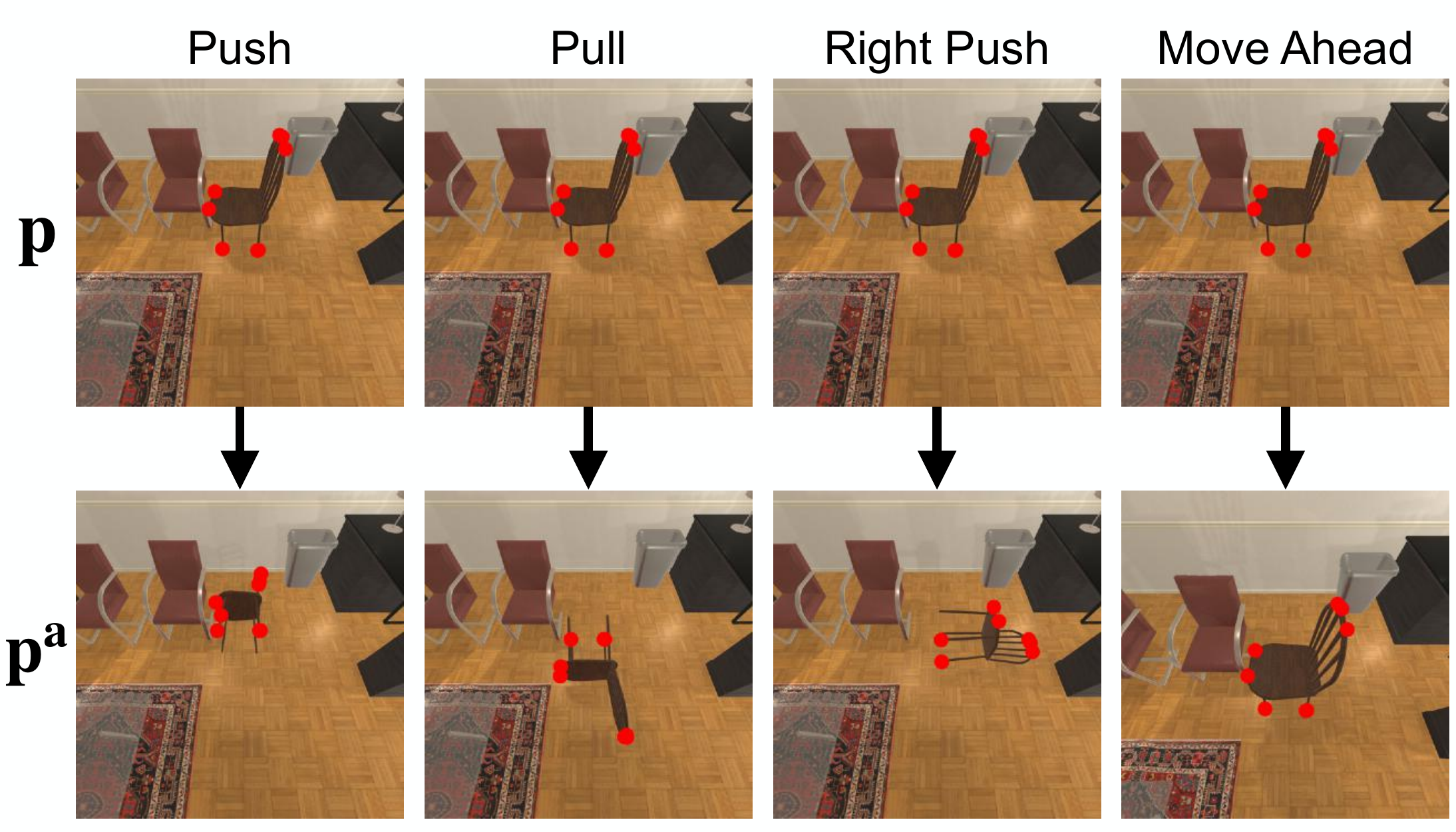}
\caption{\textbf{Keypoint examples.} The top row shows object keypoints $\textbf{p}_{o}$ and bottom row shows action-conditioned keypoints $\textbf{p}_{o}^{a}$ resulted from \texttt{Push}, \texttt{Pull}, \texttt{RightPush} and \texttt{MoveAhead} actions. The keypoints are showon in red.} \label{fig:keypoints}
\end{figure}
Our model has three main components: a visual encoder, Neural Interaction Engine, and a policy network, as illustrated in Fig.~\ref{fig:framework}. First, the visual encoder produces a representation $v$ from a visual observation $i$. The visual observation includes an RGB image captured by a mounted camera and a depth image captured by a depth sensor. The visual encoder is a convolutional neural network aiming to extract informative features from the given observation. Second, the NIE, which receives the same input observation $i$, extracts keypoints $\textbf{p}_{o}$ of an object $o \in O$, and predicts keypoint locations $\textbf{p}_{o}^{a}$ after applying each action $a \in \mathcal{A}$. 
Fig.~\ref{fig:keypoints} shows typical examples of $\textbf{p}_{\texttt{chair}}$ and $\textbf{p}_{\texttt{chair}}^{a}$ after applying \texttt{Push}, \texttt{Pull}, \texttt{RightPush} and \texttt{MoveAhead} actions. More specifically, the NIE predicts affine transformation matrices $m_{o}^{a} \in \mathbb{R}^{4\times4}$ corresponding to each object and each action. Then, we derive the $\textbf{p}_{o}^{a}$ by translating and rotating the $\textbf{p}_{o}$ via $m_{o}^{a}$ in $3$D space. Applying the affine transformation to the keypoints preserves the rigid body constraint while moving keypoints of the same object. The NIE summarizes both the extracted keypoints and the action-conditioned keypoints into an action-conditioned state feature $r^{a}$. In this way, the NIE provides possible outcomes resulting from each action to the policy network. Finally, given a goal representations $g$, the policy network utilizes both $v$ and $r^{a}$ to generate an action $a$ for the agent.

\subsection{Neural Interaction Engine}
\label{sec:NIE}

\begin{figure*}[t!]
\includegraphics[width=41pc]{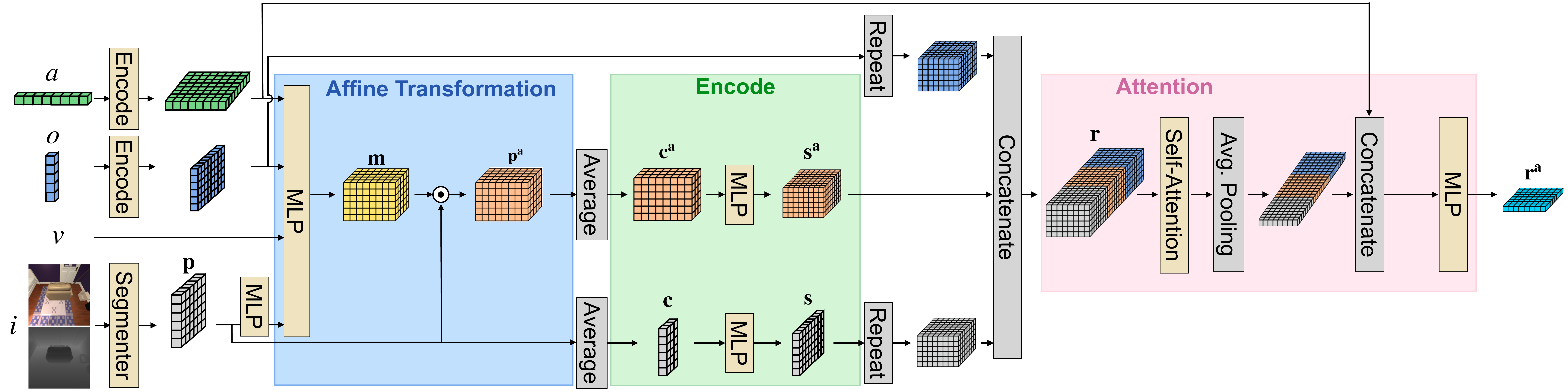}
\caption{\textbf{Neural Interaction Engine.} The inputs to the neural interaction engine are action indices, object categories, visual representation $v$ from the visual encoder, and visual observation $i$, which includes an RGB image and a depth map. After encoding each input modality, the engine uses an MLP to predict the affine transformation matrices to translate and rotate keypoints $\textbf{p}$ to $\textbf{p}^{a}$ corresponding to all objects and all actions. Then, the engine encodes the average of keypoints into hidden features $\textbf{s}$ as well as $\textbf{s}^{a}$. Finally, the engine utilizes a self-attention layer to summarize the hidden features into a semantic action-conditioned state representation $\textbf{r}^{a}$.} 
\label{fig:NIP}
\end{figure*}

The NIE operates by first extracting object keypoints $\textbf{p} \in \mathbb{R}^{O \times (N \times 3)}$, where $N$ denotes the number of keypoints, $O$ denotes the observed objects, and each $p \in \mathbb{R}^{3}$ describes a point in the three dimensional space, and then, based on these keypoints, predicting the action-conditioned keypoints $\textbf{p}^{a} \in \mathbb{R}^{O \times |\mathcal{A}| \times (N \times 3)}$ for each action $a$ in the action space $\mathcal{A}$. The engine captures a summary of possible outcomes for each action and object. These summaries are used by the policy network to sample an action $a$.

As shown in Fig.~\ref{fig:NIP}, the input to NIE includes the observation $i$, which includes an RGB frame and a depth map, the visual representation $v$ from the visual encoder, the object category embedding, and the action index embedding. The observation is first passed through a MaskRCNN~\cite{he2017mask} to obtain object segmentations. To extract the keypoints, we heuristically detect $8$ corner points in an object segment as the keypoints for this object (see Sec.~\ref{app:A} for more details). We used a heuristic approach to find the keypoints, but any other keypoint detection approach (e.g., \cite{kulkarni2019unsupervised,suwajanakorn2018discovery}) could be used instead. Further, using the depth map and camera parameters of the agent, we back project the keypoints onto the $3$-dimensional space. 

To predict the outcome of each action, the NIE predicts affine transformation matrices for each object and action, as shown in the \textit{Affine Transformation} module in Fig.~\ref{fig:NIP}. In practice, we first embed the keypoints $\textbf{p}$ into hidden features and concatenate it with the object category embedding as well as the action index embedding. Then, we use an MLP to predict the affine transformation matrix $\textbf{m} \in \mathbb{R}^{O \times |\mathcal{A}| \times 4 \times 4}$ for all objects $O$ and all actions in the action space $\mathcal{A}$. We translate and rotate the keypoints $\textbf{p}$ according to $\textbf{m}$ to obtain $\textbf{p}^{a}$. Since each $m_{o}^{a} \in \textbf{m}$ encodes the information associated with object category and the action $a$, the predicted keypoints not only contain semantic meaning, but also carry action-dependent information.

To encode keypoints and their corresponding action-conditioned keypoints, we first compute the center ($\textbf{c}$ and $\textbf{c}^{a}$) of both $\textbf{p}$ and $\textbf{p}^{a}$ by averaging the coordinates along each axis (i.e, $c_x = \frac{1}{N}\sum_{n=1}^{N} p_{x}^{n}$, $c_y = \frac{1}{N}\sum_{n=1}^{N} p_{y}^{n}$, $c_z = \frac{1}{N}\sum_{n=1}^{N} p_{z}^{n}$). Further, we employ a state encoder to encode $\textbf{c}$ and $\textbf{c}^{a}$ into hidden features ($\textbf{s}$ and $\textbf{s}^{a}$), as shown in the \textit{Encode} module in Fig.~\ref{fig:NIP}.

The hidden features $\textbf{s}$ and $\textbf{s}^{a}$ are then concatenated with the object category embedding to construct a semantic action-conditioned state representation $\textbf{r}$. Furthermore, we perform Self-Attention~\cite{vaswani2017attention} on $\textbf{r}$ over the object category axis and an Average-Pooling layer to obtain the action-conditioned state representation $\textbf{r}^{a}$, as illustrated in the \textit{Attention} module in Fig.~\ref{fig:NIP}. The reason for this step is not only to make the action-conditioned representation more compact, but also to directly associate it to each action. 

{\noindent \textbf{Integrating NIE output into the Policy Network.}} We construct a global representation $f$ by concatenating the goal representations $g$ (e.g., target location encoding for the point navigation task), visual representation $v$, and action-dependent state features $\textbf{r}^{a}$. The policy network takes $f$ as the input and outputs a probability distribution over the action space. The agent samples an action from this distribution to execute in the environment.


\subsection{Learning Objective}
\label{sec:objective}
\begin{figure}[tp]
\centering
\includegraphics[width=19pc]{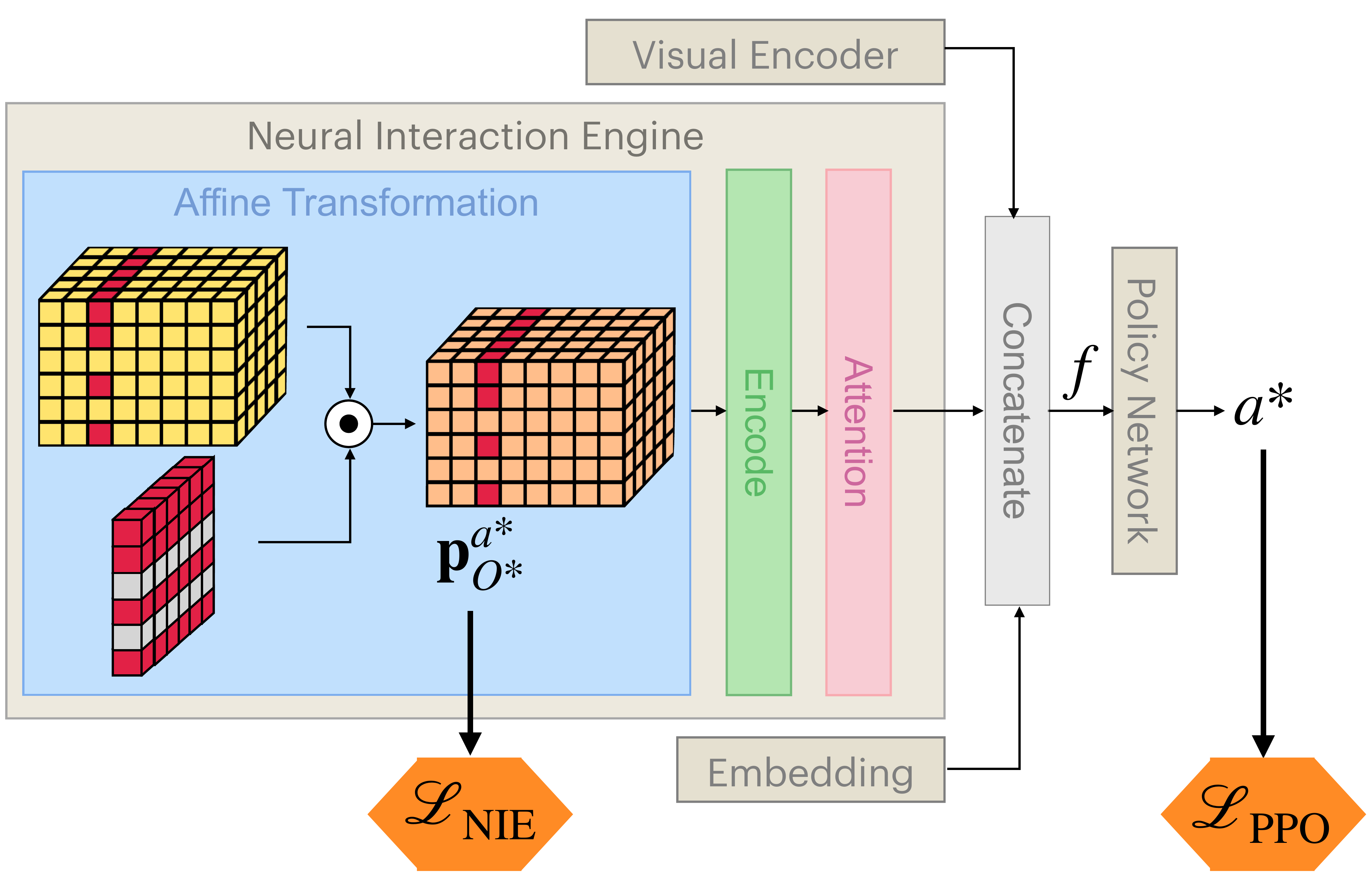}
\caption{\textbf{Training pipeline.} The entire model is trained by  $\mathcal{L}_{\text{PPO}}$ and the Affine Transformation module is trained by $\mathcal{L}_{\text{NIE}}$. However, the gradients backpropagated from $\mathcal{L}_{\text{NIE}}$ are only used to update the parameters corresponding to $\textbf{p}_{O^{*}}^{a^{*}}$, where $O^{*}$ are the observed object categories and $a^{*}$ is the action taken by the agent. The tensors corresponding to $\textbf{p}_{O^{*}}^{a^{*}}$ are highlighted in red.} \label{fig:training}
\vspace{-5mm}
\end{figure}
To train the model to learn the affine transformation matrix, we use the pose of an object before and after applying an action $a$ in the environment to construct the ground truth affine transformation matrix. Then, we apply this ground truth affine transformation matrix to the keypoints $\textbf{p}$ to obtain the ground truth action-conditioned keypoints $\textbf{t}^{a}$. We cast the learning as a regression problem and use L1 loss to optimize NIE. The agent can only pick one action to execute at each timestamp. Hence it is not possible to obtain the ground truth action-conditioned keypoints $\textbf{t}^{a}$ for all possible actions $a \in \mathcal{A}$. The agent only observes few objects among the object categories $O$, so we do not backpropagate the gradients back to the object categories that are not observed. As a result, during the training stage (as illustrated in Fig.~\ref{fig:training}), we only compute the loss for the executed action and backpropagate the gradients only through the path corresponding to $a^{*}$, the action that is actually executed by the agent and also the observed object categories $O^{*} \subset O$:

\begin{equation}
    \mathcal{L}_{\text{NIE}} = L1(\textbf{p}_{O^{*}}^{a^{*}}, \textbf{t}_{O^{*}}^{a^{*}}).
\end{equation}

\begin{figure*}[tp]
\centering
\includegraphics[width=41pc]{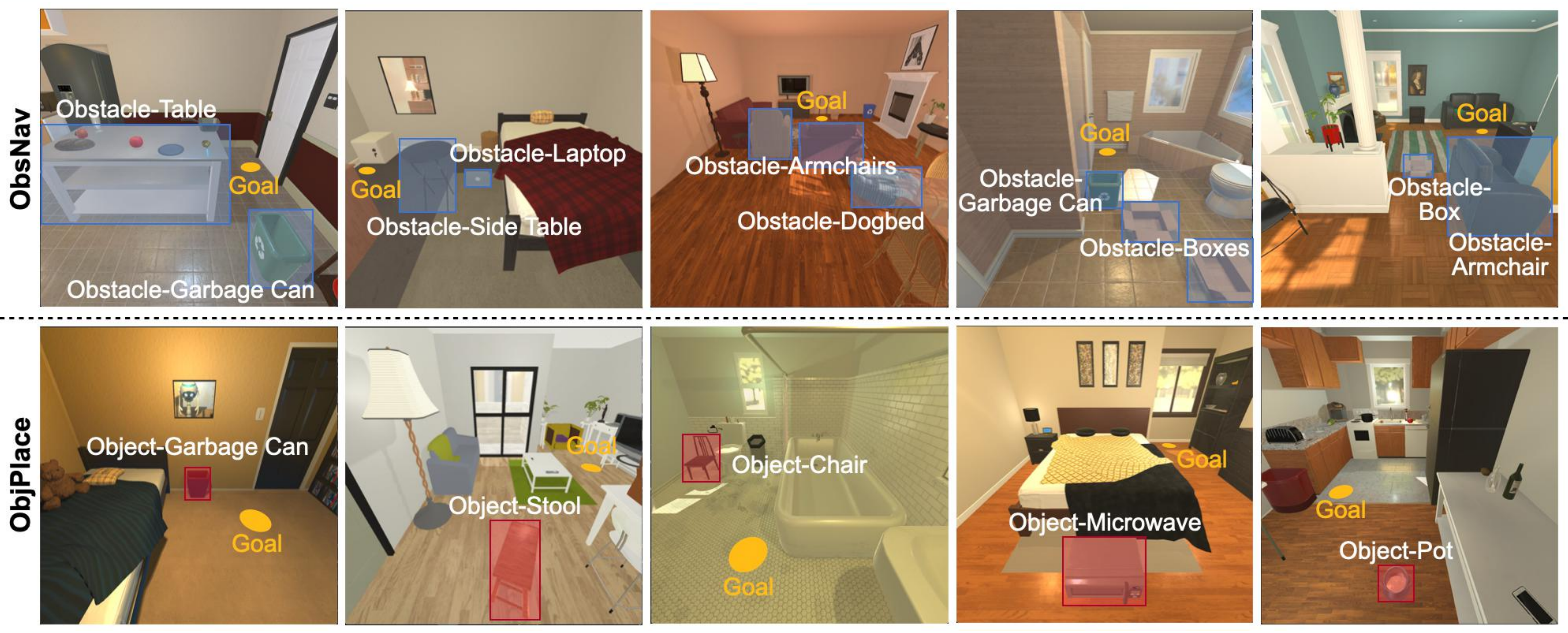}
\caption{\textbf{Dataset examples.} Top: five examples in \textit{ObsNav} dataset, where the blue boxes are obstacles and the yellow circle is the target position. Bottom: five examples in \textit{ObjPlace} dataset, where the red boxes are the object that should be displaced and the yellow circle is the target place. } \label{fig:dataset}
\vspace{-3mm}
\end{figure*}

Further, to learn the policy network, we employ the Proximal Policy Optimization (PPO)~\cite{schulman2017proximal} to perform an on-policy reinforcement learning, as illustrated in Fig.~\ref{fig:training}. The overall learning objective is $\mathcal{L} = \mathcal{L}_{\text{PPO}} + \alpha \mathcal{L}_{\text{NIE}}$,
where the $\alpha\geq 0$ is a hyperparameter controlling the relative importance of the NIE loss.

\section{Experiments}
\label{sec:exp}

To evaluate the effectiveness of the proposed Neural Interaction Engine, we evaluate it on two downstream tasks. In the following, we first describe the two downstream tasks. We then describe environment details and the datasets we have collected for training and evaluating the proposed framework. Further, we provide the implementation details in Sec.~\ref{sec:imp}. In Sec.~\ref{sec:baselines} and Sec.~\ref{sec:ablation}, we introduce our comparative baselines and variations of our model. Finally, we present quantitative and qualitative results in Sec.~\ref{sec:results}.

\noindent \textbf{Downstream tasks.} 
We consider two downstream tasks for our experiments:

\begin{itemize}
\item {\noindent \emph{ObsNav}} -- The goal of ObsNav is to move from a random starting location in a scene to specific coordinates while the path to the target point is blocked by obstacles on the floor. This is similar to PointNav \cite{anderson18} with the difference that the agent should move objects out of the way to reach the target.

\item {\noindent \emph{ObjPlace}} -- The second downstream task that we consider is ObjPlace. The goal is to move an object on the floor from a random starting location to a specified coordinate in a scene. This task requires successive application of a force to an object while navigating towards the target point.
\end{itemize}

{\noindent Successful completion of these tasks requires reasoning about the outcome of the agent actions while performing long-horizon planning. Therefore, they are suitable testbeds to evaluate our model.}

{\noindent \textbf{Environment settings.}} In this work, we perform experiments on AI2-iTHOR~\cite{kolve2017ai2} v2.7.2, which provides fairly accurate physical properties of objects. AI2-iTHOR is built using the Unity game engine which enables the simulation of physical agent-object and object-object interactions. In this environment, we consider actions $\mathcal{A}=\{$ \texttt{MoveAhead}, \texttt{RotateRight}, \texttt{RotateLeft}, \texttt{LookUp}, \texttt{LookDown}, \texttt{Push}, \texttt{Pull}, \texttt{RightPush}, \texttt{LeftPush}, \texttt{END}$\}$, where  \texttt{MoveAhead} moves the agent ahead by $0.25$ meters, \texttt{RotateRight} and \texttt{RotateLeft} change the agent's azimuth angle by $\pm90$ degrees, \texttt{LookUp} and \texttt{LookDown} rotate the agent's camera elevation angle by $\pm30$ degrees, the \texttt{Push}, \texttt{Pull}, \texttt{RightPush},  as well as \texttt{LeftPush} let the agent push (along $\pm z$ and $\pm x$ axis) the closest observed object by applying a force of $100$ newtons. The agent issues the \texttt{END} to indicate that it has completed an episode. Fig.~\ref{fig:keypoints} shows four typical examples where the agent applies \texttt{Push}, \texttt{Pull}, \texttt{RightPush}, \texttt{LeftPush} actions. Finally, we set the height and width of RGB and depth images to 224.  Thereby, the ground truth object segmentation used to learn the NIE is also of the same dimensions.

{\noindent \textbf{Data collection.}} We use \textit{Kitchens}, \textit{Living Rooms}, \textit{Bedrooms}, and \textit{Bathrooms} for our experiments (120 scenes in total). We follow the common practice for AI2-THOR wherein the first $20$ scenes are used for training, the next $5$ for validation, and the last $5$ for testing in each scene category. To collect the datasets, we use $20$ categories of objects such as \texttt{Chair}, \texttt{SideTable}, and \texttt{DogBed}. Please see Sec.~\ref{app:B} for the used objects. These objects are used as obstacles for \textit{ObsNav} and as objects that should be displaced in \textit{ObjPlace}. These objects are spawned on the floor for the downstream tasks. For each object category we have $5$ different variations. We randomly select the first $4$ variations to collect the training and validation data and use the $5$th variation to collect the test data.

To generate the dataset for \textit{ObsNav}, we utilize an undirected graph to compute the path from the agent's starting location to the target location. Then, we randomly spawn an object to block the path. To ensure that there is no way that the agent can directly reach the target location without moving an object, we repeat this process until there is no path between the agent's starting location (source node) and target location (end node). The top row in Fig.~\ref{fig:dataset} shows five examples in this dataset.

To generate the dataset for \textit{ObjPlace}, we first create a yellow mark at a random location on the floor in a scene to indicate the target location. We then spawn an object at another random location, which is at least $2$ meters away from the target location. In total, we collect $10$k training instances, $2.5$k validation instances, and $2.5$k testing instances for each task. The bottom row in Fig.~\ref{fig:dataset} shows five examples in this dataset.

\subsection{Implementation details}
\label{sec:imp}

In this work, we use the AllenAct~\cite{AllenAct} framework to conduct experiments. We train our model using both $\mathcal{L}_{\text{PPO}}$ and $\mathcal{L}_{\text{NIE}}$ simultaneously. We set the $\alpha$ parameter in Section~\ref{sec:objective} to $3$. We discuss the effect of $\alpha$ on the performance in Sec.~\ref{sec:results}. For \textit{ObsNav}/\textit{ObjPlace}, an episode is successful if the agent invokes \texttt{END} while the agent/object reaches a position within $0.2$ meters of the target position. During the training stage, we perform the on-policy reinforcement learning (PPO) with $80$ processes simultaneously. We use Adam with initial learning rate of $3\cdot 10^{-4}$ which decays linearly to 0 during training. We set the standard RL reward discounting parameter $\gamma$ to $0.99$, $\lambda_{\text{gae}}$ to $0.95$, and number of update steps to $30$ for $\mathcal{L}_{\text{PPO}}$. The gradients $\Delta$ are clipped to satisfy $|\Delta| <= 0.5$. We train the policy for $10$ million steps and evaluate the model every $1$ million steps. 

During the training stage, we use the ground truth object mask provided by the environment, while in the testing stage, we employ a pre-trained MaskRCNN \cite{he2017mask} to extract the segmentation. The number of output classes for both ground truth segmentation and MaskRCNN is $21$, including $20$ used objects and a background class. We use~\cite{wu2019detectron2} to pre-train the MaskRCNN (ResNet-50 with FPN) on our training scenes with $8$k images for $10$ epochs. Please see Sec.~\ref{app:C} for more details about the qualitative results generated by the MaskRCNN on our validation scenes. 

{\noindent \textbf{Model architecture.}} Because the visual observation $i$ includes an RGB image and a depth image, we employ two different CNNs, with different input number of channels, in the Visual Encoder to handle these two observations separately. After the CNNs extract features from both observations, we use a linear layer to fuse the two features together. In both tasks, we provide the observation from a GPS sensor to the policy network. The GPS's observation is a coordinate of the target position for \textit{ObsNav} or the target place for \textit{ObjPlace}. In addition to the GPS's observation, we employ a look-up embedding to encode the category of the target object for \textit{ObjPlace}. The Encode and MLP shown in Fig.~\ref{fig:NIP} are a look-up embedding layer and a multi-layer perceptron, respectively. The Self-Attention layer has three MLPs as well to handle the key, query, and value embedding. Our policy network consists of a GRU state encoder, a linear layer for the actor (policy), and a linear layer for the critic (value). Please refer to Sec.~\ref{app:D} for more details about each model components such as the number of layers and hidden dimension.

{\noindent \textbf{Reward shaping.}}  We consider a task successful if the agent invokes the \texttt{END} when the agent achieves the goal. For \textit{ObsNav}, the goal is to reach within a certain distance ($0.2$ meters) to the target location and for \textit{ObjPlace}, the object should have overlap with the yellow target mark. If the agent succeeds in an episode, we provide a reward of $+10$. We find reward shaping~\cite{ng1999policy} important to learn the policy in the two studied tasks. We implement reward shaping for each task as follows:

{\noindent$\bullet$ \textit{ObsNav}}: Similar to~\cite{Savva_2019_ICCV}, we implement the reward shaping based on geodesic distance. We provide a reward to the agent after it takes an action based on the change in the geodesic distance between the current agent position and the target position. If the agent takes an action resulting in a decrease of the geodesic distance, the agent receives the decreased amount as the reward. Otherwise, if the taken action causing an increase in the geodesic distance, the agent receives the amount of increase as a penalty. There are obstacles blocking the paths to the destination, so we also encourage the agent to take actions to move the obstacles out of the way. Therefore, the environment provides a $-0.5$ penalty if the agent takes any action that blocks a path between the agent and the goal and conversely, a $0.5$ reward if the agent's action opens a new path to the goal. We let $r_{\text{dis/appear}}=-0.5$ if the agent's action resulted in blocking a path, $r_{\text{dis/appear}}=0.5$ if the agent's action opened a path, and $r_{\text{dis/appear}}=0$ otherwise.

{\noindent$\bullet$ \textit{ObjPlace}}: For this task, we perform reward shaping only based on the geodesic distance. We provide a reward to the agent after it takes an action according to the change in the geodesic distance between the current object position and the target position. 

To encourage the agent to finish the task as quickly as possible, we also add a small penalty $-0.01$ at each step. As a result, the total reward at step $t$ is:
\begin{gather}
\nonumber r_{t} = \begin{cases}r_{s} + r_{\text{dis/appear}} + d_{t-1} - d_{t} + p& \text{if goal is reached,} \\
r_{\text{dis/appear}} + d_{t-1} - d_{t} + p& \text{otherwise,}
\end{cases}
\end{gather}
where $r_{s}$ is set to $10$, $d_{t}$ denotes the geodesic distance between the agent (object) and the target position at step $t$, and $p$ is the step penalty which equals $-0.01$ and $-0.002$ for \textit{ObsNav} and \textit{ObjPlace}, respectively. Note that the $r_{\text{dis/appear}}$ is removed in the \textit{ObjPlace} task. However, adding $r_{\text{dis/appear}}$ is essential to the \textit{ObsNav} task, since the policy network with visual observation $i$ but without $r_{\text{dis/appear}}$ does not generalize to the unseen environments even after $10$ million steps of training.

\subsection{Baselines}
\label{sec:baselines}

We compare our model with the following baseline methods. Each baseline uses the same visual encoder and policy network unless stated otherwise.

{\noindent \textbf{PPO}.} This baseline is a Reinforcement Learning based approach that has a visual encoder to extract features from visual observation $i$ and an embedding layer to encode the GPS readings. The model is trained by Proximal Policy Optimization~\cite{schulman2017proximal} and we use the same learning hyparameters mentioned in Sec.~\ref{sec:imp} to train using this method.

{\noindent \textbf{RGB-D and object segmentation input (RGB-D-S)}.} This baseline includes a segmentation image as well as the visual observation $i$. We extend the Visual Encoder by another CNN to extract features from the segmentation image. As mentioned in Sec.~\ref{sec:imp}, we use the ground truth segmentation during the training stage and the results generated by MaskRCNN (fine-tuned on our data) during the evaluation.

{\noindent \textbf{RGB-D and keypoints input (RGB-D-K).}} To understand if the keypoints representation extracted from object segmentation is more meaningful than a pure segmentation image input, we implement this baseline by including the keypoints extracted by the same heuristic corner detector (Sec.~\ref{sec:NIE}) used in our model as an additional input. To encode the keypoints, we use the same model architecture as the NIE module to obtain the semantic action-conditioned state representation $\textbf{r}^{a}$ as well. However, the $\mathcal{L}_{\text{NIE}}$ is not used to learn the NIE module in this baseline. The parameters are updated by the gradients from $\mathcal{L}_{\text{PPO}}$ only.

{\noindent \textbf{PPO + auxiliary loss.}} We implement a baseline based on CPC$|$A~\cite{guo2018neural} to facilitate the policy learning upon the PPO baseline. During the training stage, the CPC$|$A utilizes Contrastive Predictive Coding as an auxiliary loss to perform predictive representation learning. To have a fair comparison, we use a GRU with the same hidden size and only predict one time step in the future.

\subsection{Ablations}
\label{sec:ablation}

To perform ablation studies, we evaluate the following variations of our NIE model.

\noindent \textbf{NIE w/o visual observations.} To understand if the visual observation $i$ can help the prediction of affine transformation matrices and the action-conditioned keypoints $\textbf{p}^{a}$, we implement this model by removing the visual input from the NIE. Therefore, the NIE only takes the keypoints in coordinate representation with action indices as well as object categories. We use the same hyperparameters and optimization approach mentioned in Sec.~\ref{sec:imp} to train this model.

\noindent \textbf{NIE w/ $1 \times \mathcal{L}_{\text{NIE}}$.} We decrease $\alpha$, which is used to balance the $\mathcal{L}_{\text{NIE}}$ and $\mathcal{L}_{\text{PPO}}$. This provides us with an insight about the importance of $\mathcal{L}_{\text{NIE}}$ to learn the entire model.

\noindent \textbf{NIE w/ $10 \times \mathcal{L}_{\text{NIE}}$.} In this ablation study, we increase the $\alpha$, which is used to balance the $\mathcal{L}_{\text{NIE}}$ and $\mathcal{L}_{\text{PPO}}$, to $10$. This study shows if a large value of $\alpha$ would have a negative impact on the final performance.

\begin{figure*}[tp]
\centering
\includegraphics[width=41pc]{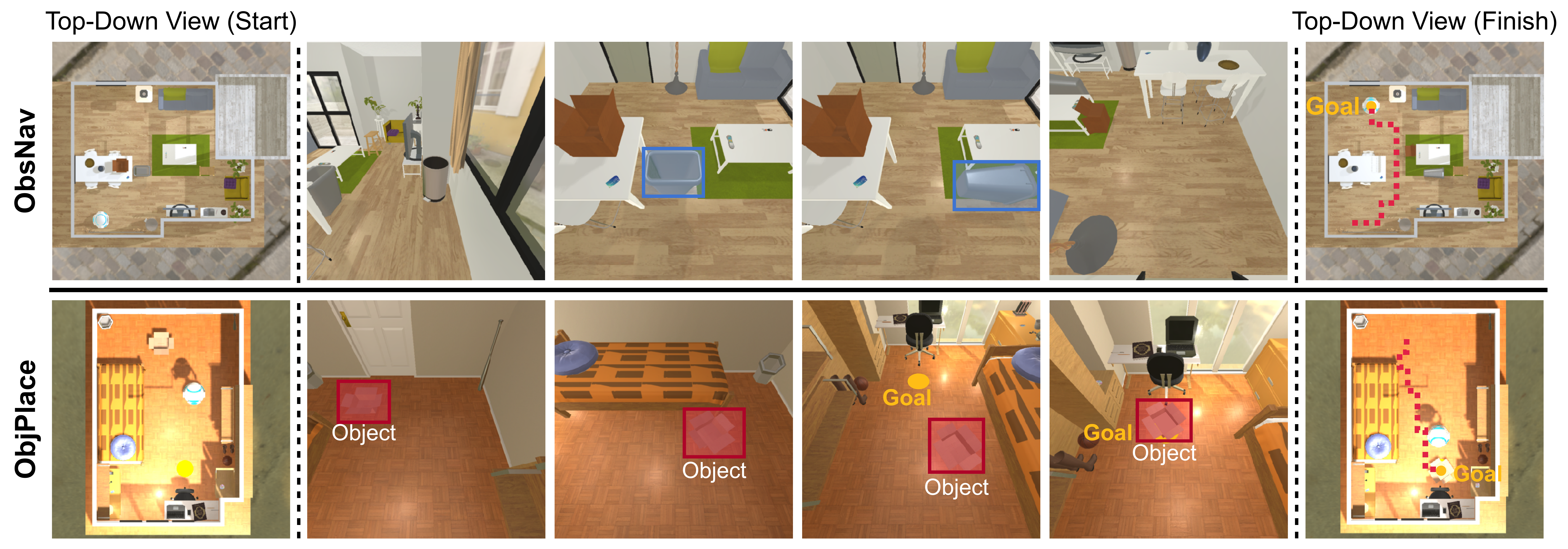}
\caption{\textbf{Qualitative results.} Top: An example of the \textit{ObsNav} task is shown. The blue box is the obstacle the agent should move away to unblock the path (the blue marking is just for visualization purposes and not visible to the agent). The agent's movement is shown by a dashed trajectory in red in the rightmost image. Bottom: An example of the \textit{ObjPlace} task, where the red box is the object that should be displaced and the orange circle is the target location. The object's movement is shown by a trajectory in red color.}
\vspace{-0.4cm}
\label{fig:qualitative}
\end{figure*}

\subsection{Results}
\label{sec:results}

\noindent \textbf{Evaluation Metrics.} We evaluate all models by \textit{Success Rate (SR)}, \textit{Final Distance to Target (FDT)}, and \textit{Success weighted by Path Length (SPL)}~\cite{anderson2018evaluation} for both tasks. \textit{SR} is the ratio of the number of successful episodes to the total number of episodes, \textit{FDT} is the average distance between agent/object and the target position as the agent issues \texttt{END} or an episode reaches the maximum number of allowed steps ($500$), and the \textit{SPL} is defined as $\frac{1}{N}\sum^{N}_{n=1}S_{n}\frac{L_{n}}{max(P_{n}, L_{n})}$, where $N$ is the number of episodes, $S_{n}$ denotes a binary indicator of success in the episode $n$, $P_{n}$ is the path length, and $L_{n}$ is the shortest path distance in episode $n$.

\begin{table}[t!]
\begin{center}
\begin{tabular}{llll}
\bf Methods & \bf SR (\%) $\uparrow$& \textbf{FDT} (m) $\downarrow$ & \bf SPL $\uparrow$ \\ \hline
\multicolumn{4}{l}{\bf Baselines:} \\
\,\,\,PPO~\cite{schulman2017proximal} & 67.1 & 0.605 & 25.7  \\
\,\,\,RGB-D-S & 62.8 & 0.499 & 25.0 \\
\,\,\,RGB-D-K & 70.9 & 0.459 & 25.8  \\
\,\,\,CPC$|$A~\cite{guo2018neural} & 73.8 & 0.370 & 29.8  \\
\hline
NIE {\bf (ours)} & \textbf{80.0} & 0.304 & \textbf{31.3} \\
\hline
\multicolumn{4}{l}{\bf Ablations:} \\
\,\,\,NIE w/o VO & 72.7 & 0.375 & 29.2  \\
\,\,\,NIE w/ $1 \times \mathcal{L}_{\rm{NIE}}$ & 74.1 & 0.377 & 29.7  \\
\,\,\,NIE w/ $10 \times \mathcal{L}_{\rm{NIE}}$ & 78.2 & \textbf{0.278} & 31.0  \\
\hline
\end{tabular}
\end{center}
\vspace{-5mm}
\caption{\textbf{ObsNav results.} We show the result of our method (referred to as `NIE') along with baselines and ablations of our model. We use $\uparrow$ and $\downarrow$ to denote if larger or smaller values are preferred. We repeat the experiments three times and report the average.}
\label{tab:obsnav}
\vspace{-4mm}
\end{table}

\noindent \textbf{ObsNav.} The quantitative results of the \emph{ObsNav} task are shown in Table~\ref{tab:obsnav}. Our method outperforms the baselines in all three metrics, which justifies the effect of using the NIE model. The performance drops for `NIE w/o VO' ablations, which shows that visual information is required to estimate the location of objects. For example, if an object is pushed against a wall, the visual information helps to reason that the object will not move. It is not feasible to make such predictions just by using the keypoint information alone. Our results on `NIE w/ $1 \times \mathcal{L}_{\rm{NIE}}$' and `NIE w/ $10 \times \mathcal{L}_{\rm{NIE}}$' show that completely relying on the NIE model is not sufficient and we need exploration as well. On the other hand, exploration alone is not sufficient. Therefore, a good balance between future prediction and exploration is required. 

\begin{table}[t!]
\begin{center}
\begin{tabular}{llll}
\bf Methods & \bf SR (\%) $\uparrow$& \textbf{FDT} (m) $\downarrow$ & \bf SPL $\uparrow$ \\ \hline
\multicolumn{4}{l}{\bf Baselines:} \\
\,\,\,PPO~\cite{schulman2017proximal} & 1.2 & 3.18 & 0.85  \\
\,\,\,RGB-D-S & 1.2 & 3.15 & 0.85  \\
\,\,\,RGB-D-K & 1.3 & 2.84 & 0.88  \\
\,\,\,CPC$|$A~\cite{guo2018neural} & 12.0 & 2.35 & 9.3  \\
\hline
NIE {\bf (ours)} & \textbf{17.5} & 2.22 & \textbf{14.2} \\
\hline
\multicolumn{4}{l}{\bf Ablations:} \\
\,\,\,NIE w/o VO & 0.8 & 3.07 & 0.41  \\
\,\,\,NIE w/ $1 \times \mathcal{L}_{\rm{NIE}}$ & 15.3 & \textbf{2.11} & 13.1  \\
\,\,\,NIE w/ $10 \times \mathcal{L}_{\rm{NIE}}$ & 13.6 & 2.26 & 11.5  \\
\hline
\end{tabular}
\end{center}
\vspace{-5mm}
\caption{\textbf{ObjPlace results.} We show the result of our method (referred to as `NIE') along with baselines and ablations of our model. We use $\uparrow$ and $\downarrow$ to denote if larger or smaller values are preferred. We repeat the experiments three times and report the average.}
\label{tab:obsplace}
\vspace{-5mm}
\end{table}

\noindent \textbf{ObjPlace.} The results are shown in Table~\ref{tab:obsplace}. As shown, there is a huge difference between the baseline models and our model. We investigated the reason for this huge gap. Most of the time the baseline agent pushes other objects as well and eventually blocks the path towards the target. 

\noindent \textbf{Qualitative Results.} We show qualitative results in Fig.~\ref{fig:qualitative}. The top row shows a successful episode of ObsNav, where the agent pushes the garbage can away to unblock the path. The bottom row show an example of the ObjPlace task, where the agent moves the box toward the goal position. It is interesting to note that the agent goes around the object of interest so it can push it towards the target location. We provide a supplementary video\footnote{\url{https://youtu.be/GvTI5XCMvPw}} to show more successful as well as failure cases, and qualitative results of future keypoints prediction conditioned on the actions in Sec.~\ref{app:E}.

\vspace{-0.2cm}
\section{Conclusion}
\vspace{-0.2cm}
We study the problem of predicting the outcome of actions in the context of embodied visual navigation tasks. We propose Neural Interaction Engine (NIE) to encode the changes to the environment caused by navigation and interaction actions of the agents. We incorporate NIE into a policy network and show its effectiveness in two downstream tasks that require long-horizon planning. The goal of the first task is to reach a target point in an environment while the paths to the target are blocked. The second task requires navigating to a target point while pushing an object. Our evaluations show the effectiveness of the NIE model in both scenarios, where we achieve significant improvements over the methods without the capability of predicting the effect of actions on the surrounding environment.

\subsection*{Acknowledgements.}

We thank members from RAIVN Lab at the University of Washington and PRIOR team at Allen Institute for AI for valuable feedbacks on early versions of this project. This work is in part supported by NSF IIS 1652052, IIS 17303166, DARPA N66001-19-2-4031, DARPA W911NF-15-1-0543 and gifts from Allen Institute for AI.

{\small
\bibliographystyle{ieee_fullname}
\bibliography{egbib}
}

\clearpage
\vfill

\appendix


\section{Heuristic corner detector}\label{app:A}
Fig.~\ref{fig:keypoints_detector} (a) shows our heuristic keypoints detector pipeline. More specifically, we first use an object segmentation model to obtain the segmentation $M$ corresponding to object $o=$ \texttt{GarbageCan}. Then, we apply a heuristic corner detector to detect $8$ corner points. Note that we use the ground truth segmentation of each object in the training stage, while in the testing stage, we utilize a pretrained MaskRCNN (Sec.~\ref{app:C}) to obtain the object segmentation.
Further we present the details of our heuristic corner detector in Fig.~\ref{fig:keypoints_detector} (b), where the $8$ corner points are obtained by $8$ different criteria and each of the points has to be inside the segmentation $M$:

\begin{gather}
    \nonumber
    p_{1} = \underset{x,\ p_{x, y} \in M}{max}\ p_{x,y} \\
    \nonumber
    p_{2} = \underset{y,\ p_{x, y} \in M}{max}\ p_{x,y} \\
    \nonumber
    p_{3} = \underset{x,\ p_{x, y} \in M}{min}\ p_{x,y} \\
    \nonumber
    p_{4} = \underset{y,\ p_{x, y} \in M}{min}\ p_{x,y} \\
    \nonumber
    p_{5} = \underset{x + y,\ p_{x, y} \in M}{max}\ p_{x,y} \\
    \nonumber
    p_{6} = \underset{x + y,\ p_{x, y} \in M}{min}\ p_{x,y} \\
    \nonumber
    p_{7} = \underset{x - y,\ p_{x, y} \in M}{max}\ p_{x,y} \\
    \nonumber
    p_{8} = \underset{x - y,\ p_{x, y} \in M}{min}\ p_{x,y},
\end{gather}

where $(x, y)$ is an image coordinate, and $M$ denotes the object segmentation. Based on this heuristic corner detector, we are able to get reliable keypoints from object segmentation. More keypoint examples obtained by our heuristic keypoints detector are shown in Fig.~\ref{fig:keypoints_detector_example}.

\begin{center}
    \captionsetup{type=figure}
	\footnotesize
	\includegraphics[width=19pc]{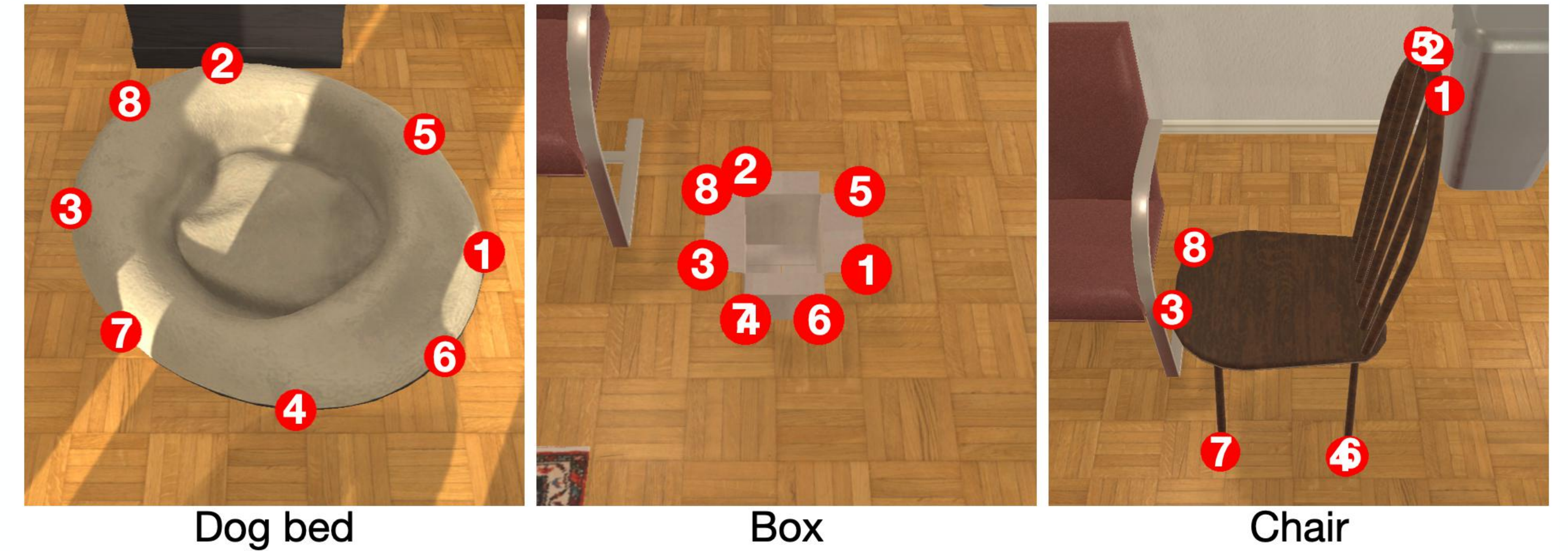}
	\captionof{figure}{\textbf{Keypoints examples.} Examples keypoints obtained by our keypoint detector.
	}
	\label{fig:keypoints_detector_example}
\end{center}
\begin{center}
    \captionsetup{type=figure}
	\footnotesize
	\includegraphics[width=19pc]{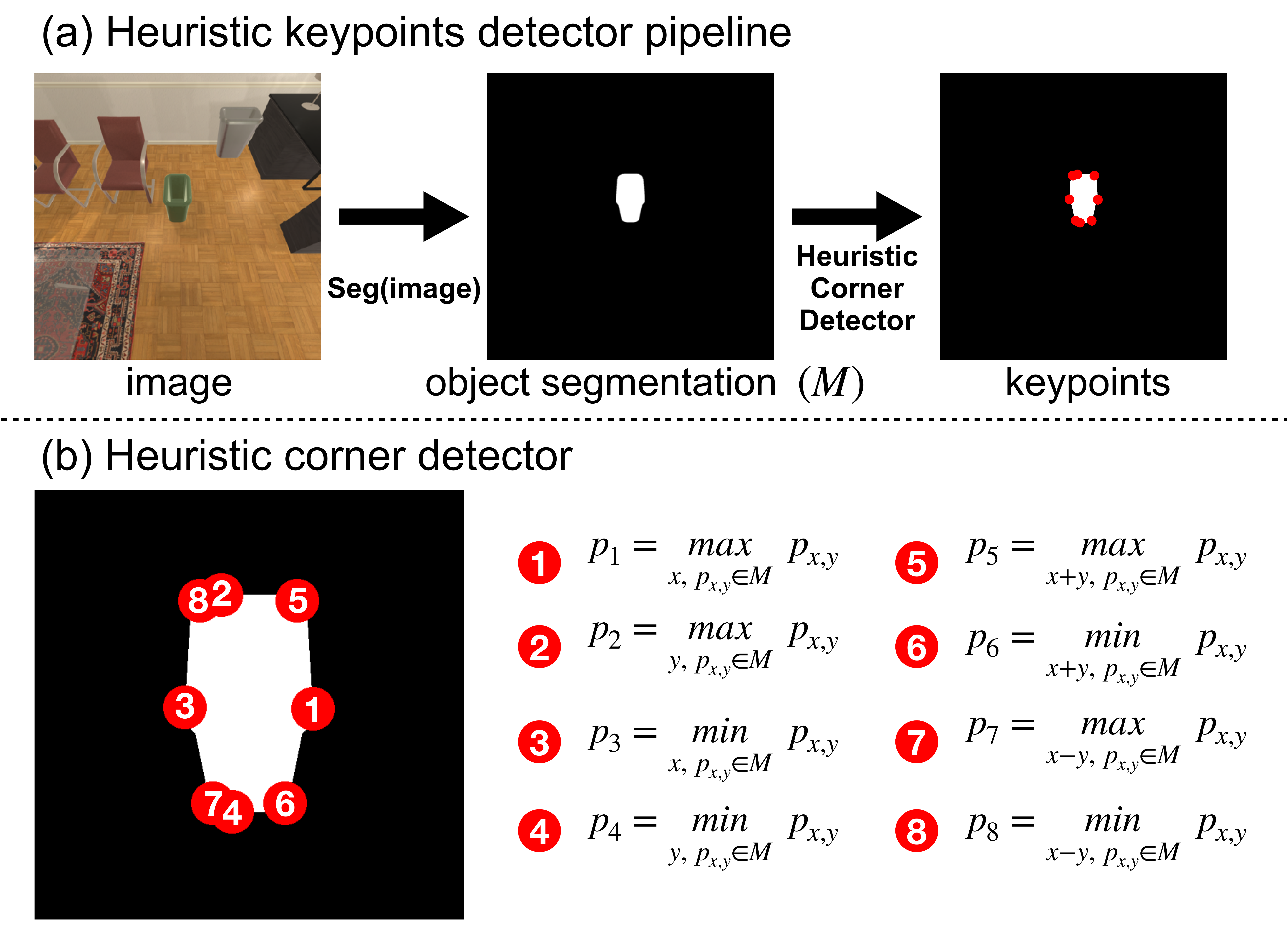}
	\captionof{figure}{\textbf{Keypoint detector details.} (a) Heuristic keypoint detector pipeline. (b) Heurisitc corner detector.
	}
	\label{fig:keypoints_detector}
\end{center}

\section{Complete list of objects}\label{app:B}
\vspace{-2mm}
We use $20$ objects for the experiments: \textit{alarm clock}, \textit{apple}, \textit{armchair}, \textit{box}, \textit{bread}, \textit{chair}, \textit{desk}, \textit{dining table}, \textit{dog bed}, \textit{garbage can}, \textit{laptop}, \textit{lettuce}, \textit{microwave}, \textit{pillow}, \textit{pot}, \textit{side table}, \textit{sofa}, \textit{stool}, \textit{television} and \textit{tomato}.

\section{MaskRCNN results}\label{app:C}
\vspace{-2mm}
We evaluate our pretrained MaskRCNN (ResNet-50 with FPN) on our testing scenes with $\approx 2$k images. The checkpoint at the $10$th epoch achieves $47.4\text{AP}$ and $64.3\text{AP}_{50}$. Fig.~\ref{fig:mask_rcnn_results} shows qualitative results on $20$ used objects in one of the testing scenes \texttt{LivingRoom227}.

\section{Details of the model architecture}\label{app:D}

Fig.\ref{fig:NIP_details} and Fig.~\ref{fig:overview_details} summarize the details of the architecture for visual encoder, goal embedding, policy network, and NIE model.

\begin{center}
    \captionsetup{type=figure}
	\footnotesize
	\includegraphics[width=20pc]{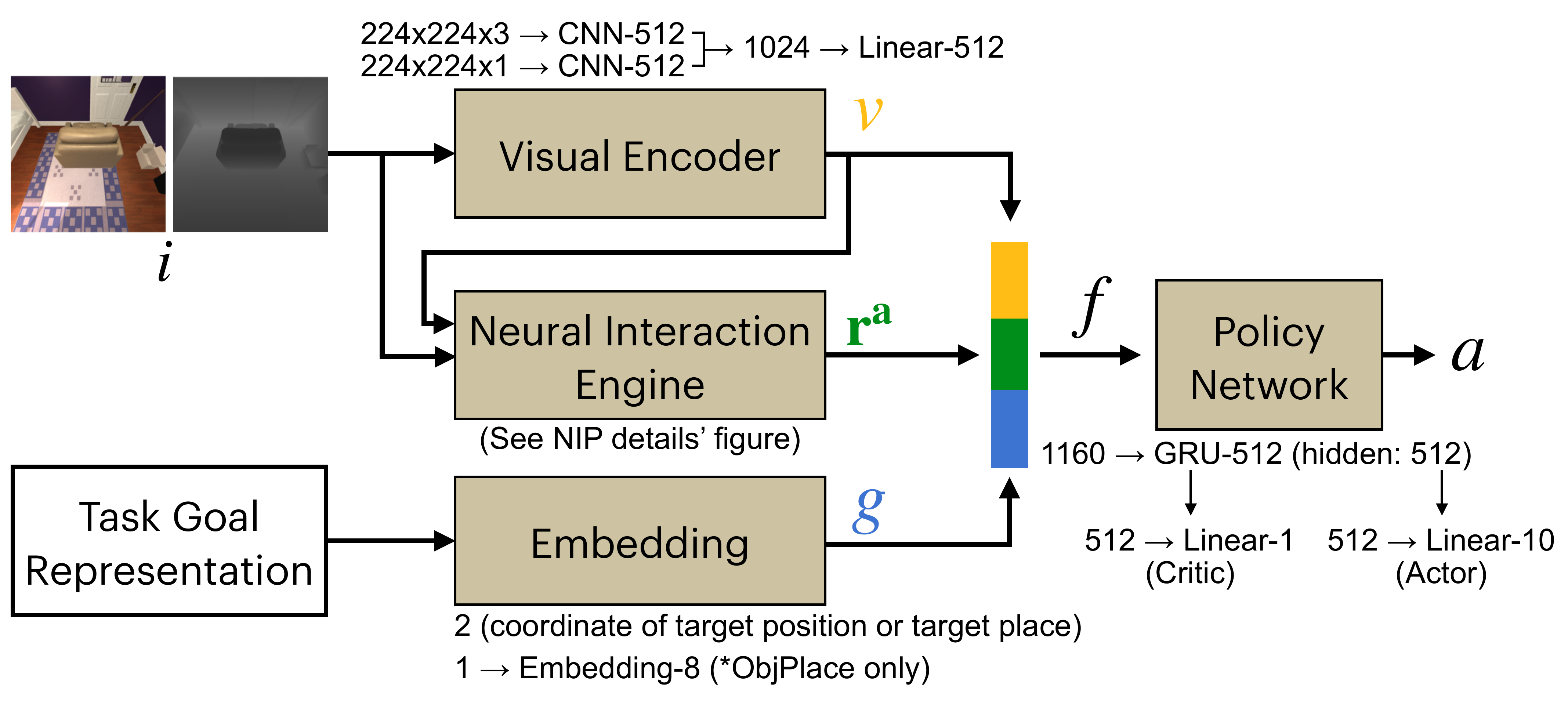}
	\captionof{figure}{\textbf{Detailed architecture of the visual encoder, goal embedding, and policy network.}
	}
	\label{fig:overview_details}
\end{center}

\twocolumn[{%
\renewcommand\twocolumn[1][]{#1}
\begin{center}
    \captionsetup{type=figure}
	\footnotesize
	\includegraphics[width=41pc]{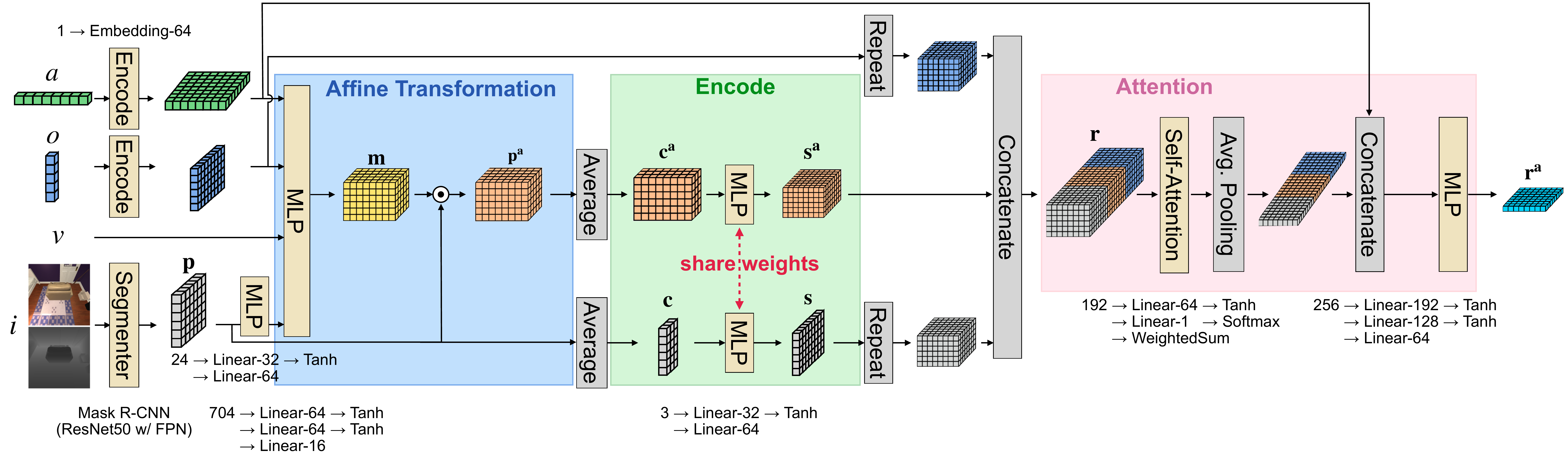}
	\captionof{figure}{\textbf{Detailed architecture of the NIE model.}
	}
	\label{fig:NIP_details}
\end{center}
}]

\section{Action-conditioned keypoints $\textbf{p}^{a}$ results}\label{app:E}

We evaluate our action-conditioned keypoints $\textbf{p}^{a}$ prediction on the testing set. Our model achieves $0.148$ and $0.114$ L1 loss estimation over $8$ keypoints on the \textit{ObsNav} and \textit{ObjPlace}, respectively. We found the model performs worse in the \textit{ObsNav} because there are more objects (e.g., obstacles) in this task.
Fig.~\ref{fig:keypoints_prediction_results} shows the qualitative results of our action-conditioned keypoint prediction.

\twocolumn[{%
\begin{center}
    \captionsetup{type=figure}
	\footnotesize
	\includegraphics[width=33pc]{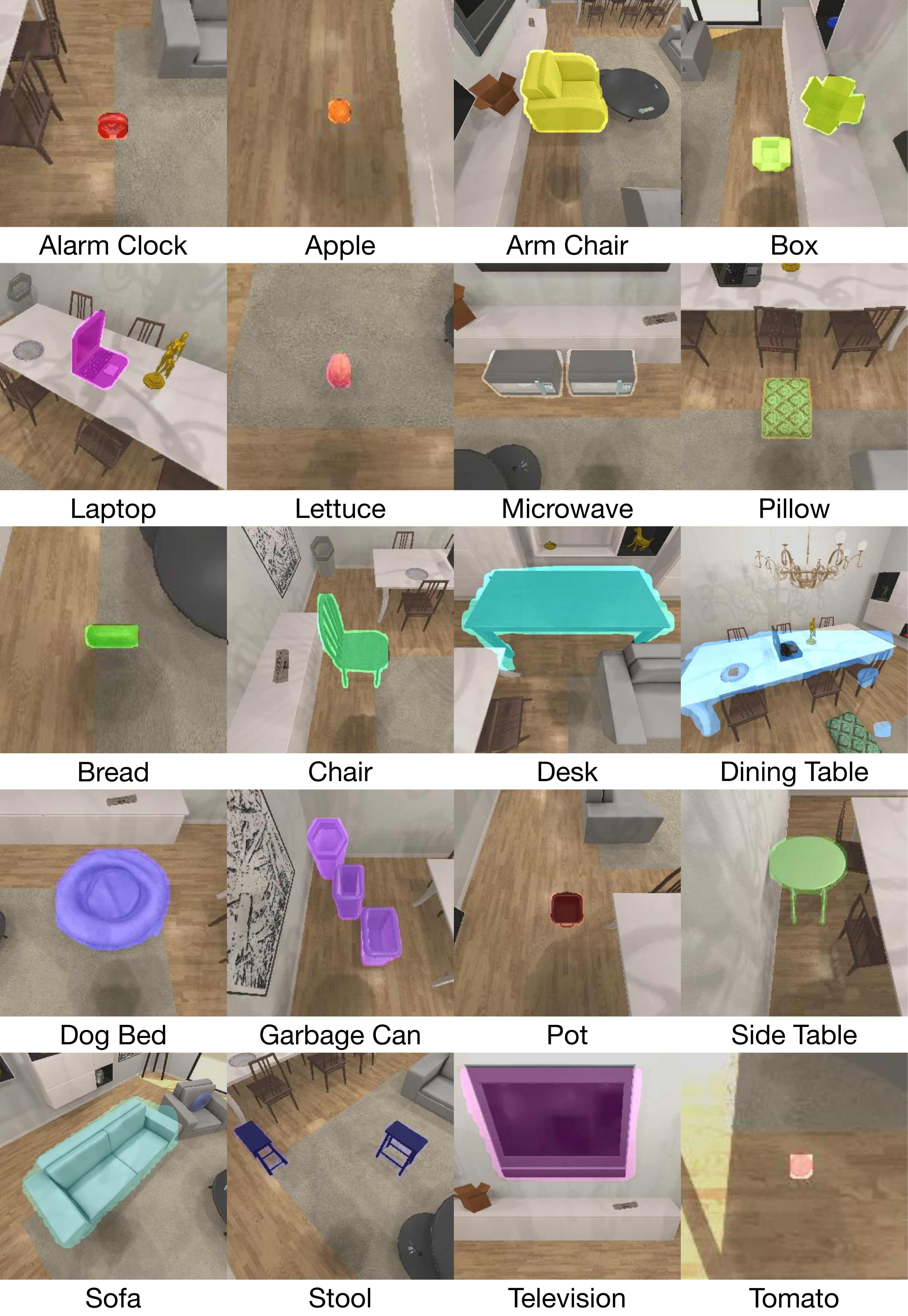}
	\captionof{figure}{\textbf{MaskRCNN's qualitative results on $20$ used objects.} We randomly spawn $20$ objects in the testing scene \texttt{LivingRoom227} and apply the pretrained MaskRCNN to obtain the segmentation results. The object prediction score is set to $0.5$ and the segmentation probability is set to $0.1$.
	}
	\label{fig:mask_rcnn_results}
\end{center}
}]

\twocolumn[{%
\renewcommand\twocolumn[1][]{#1}
\begin{center}
    \captionsetup{type=figure}
	\footnotesize
	\includegraphics[width=39pc]{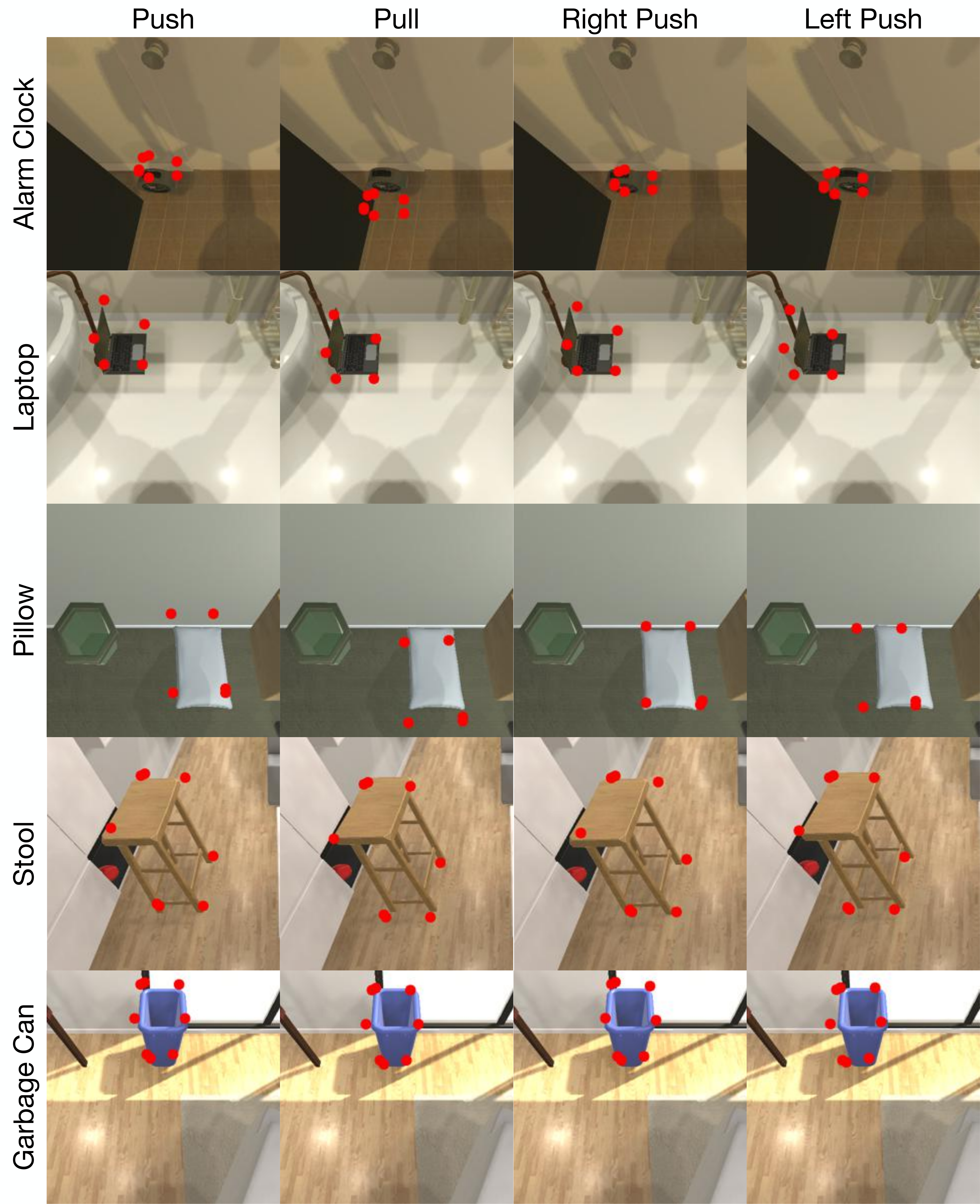}
	\captionof{figure}{\textbf{Qualitative results of action-conditioned keypoints $\textbf{p}^{a}$ prediction.} We show our action-conditioned keypoints $\textbf{p}^{a}$ prediction results over $4$ actions on $5$ objects in $4$ different testing scene (from top to bottom: \texttt{Kitchen27}, \texttt{Bathroom430}, \texttt{Bedroom328}, and \texttt{LivingRoom227}). The predicted keypoints are shown in red color.
	}
	\label{fig:keypoints_prediction_results}
\end{center}
}]

\end{document}